\documentclass[letterpaper, 10 pt, conference]{ieeeconf}
\UseRawInputEncoding

\usepackage{amsmath,amsfonts}
\usepackage{array}
\usepackage[caption=false,font=normalsize,labelfont=sf,textfont=sf]{subfig}
\usepackage{textcomp}
\usepackage{stfloats}
\usepackage{url}
\usepackage{verbatim}
\usepackage{graphicx}
\usepackage{cite}

\usepackage{xcolor}

\usepackage[ruled,vlined]{algorithm2e}
\usepackage{bm}

\usepackage{balance}

\IEEEoverridecommandlockouts                          
\begin{document}
	
	\title{\LARGE \bf
		Point-Wise Vibration Pattern Production via a Sparse Actuator Array for Surface Tactile Feedback}
	
	\author{Xiaosa Li$^{\dag}$, Runze Zhao$^{\dag}$, Chengyue Lu, Xiao Xiao and Wenbo Ding*
		\thanks{$^{\dag}$These authors contributes equally to this article.}
		\thanks{X.~Li, R.~Zhao, C.~Lu, X.~Xiao, and W.~Ding are with Tsinghua-Berkeley Shenzhen Institute, Tsinghua Shenzhen International Graduate School, Tsinghua University, China. W. Ding is the corresponding author. E-mail: (\{lixs21, zhaorz22, xiaox22\}@mails.tsinghua.edu.cn, \{luchengyue, ding.wenbo\}@sz.tsinghua.edu.cn). W.~Ding is also with RISC-V International Open Source Laboratory, Shenzhen, China, 518055. \textit{(Corresponding author: Wenbo Ding)}}
        \thanks{This article has supplementary materials of hardware and software available at https://github.com/Lixiaosa1996/VibrationPattern.git.}
	}
	
	\maketitle
	
	\begin{abstract}
		Surface vibration tactile feedback is capable of conveying various semantic information to humans via the handheld electronic devices, like smartphone, touch panel, and game controller. However, covering the whole device contacting surface with dense actuator arrangement can affect its normal use, how to produce desired vibration patterns at any contact point with only several sparse actuators deployed on the handled device surface remains a significant challenge. In this work, we develop a tactile feedback board with only five actuators in the size of a smartphone, and achieve the precise vibration pattern production that can focus at any desired position all over the board. Specifically, we investigate the vibration characteristics of single passive coil actuator, and construct its vibration pattern model at any position on the feedback board surface. Optimal phase and amplitude modulation, found with the simulated annealing algorithm, is employed with five actuators in a sparse array. And all actuators' vibration patterns are superimposed linearly to synthetically generate different onboard vibration energy distribution for tactile sensing. Experiments demonstrated that for point-wise vibration pattern production on our tactile board achieved an average level of about 0.9 in the Structural Similarity Index Measure (SSIM) evaluation, when compared to the ideal single-point-focused target vibration pattern. The sparse actuator array can be easily embedded into usual handheld electronic devices, which shows a good significant implication for enriching their haptic interaction functionalities.
	\end{abstract}
	

	\section{Introduction}
	
	Tactile feedback has been adopted to compensate for instability in various human-machine interaction applications~\cite{rodriguez2021unstable}. Vibration is the most common form for tactile feedback, due to its advantages including noticeable effects, simple generation, easy deployment, and resilience to environmental interference~\cite{sklar1999good,jimenez2014evaluation}. Incorporating vibration pattern feedback into handheld devices, like smartphone, touch panel and game controller, can help to achieve precise manipulation~\cite{croce2016enhancing}, as well as to enhance the interaction immersion~\cite{pamungkas2016electro}. And it has held promising applications in social and personal engagement~\cite{ding2023surface}, touch control and feedback~\cite{chen2022haptag}, and gaming and entertainment~\cite{han2017boes}. However, for these handheld devices, it is challenging to deploy a dense actuator array like in~\cite{yu2019skin,yao2022encoding,guo2021mrs,guo2022electromagnetic} to cover the whole surface, which will perform a devastating effect on the device manipulation. 

	\begin{figure}[htbp]
	\centering
	\includegraphics[width=7.5cm]{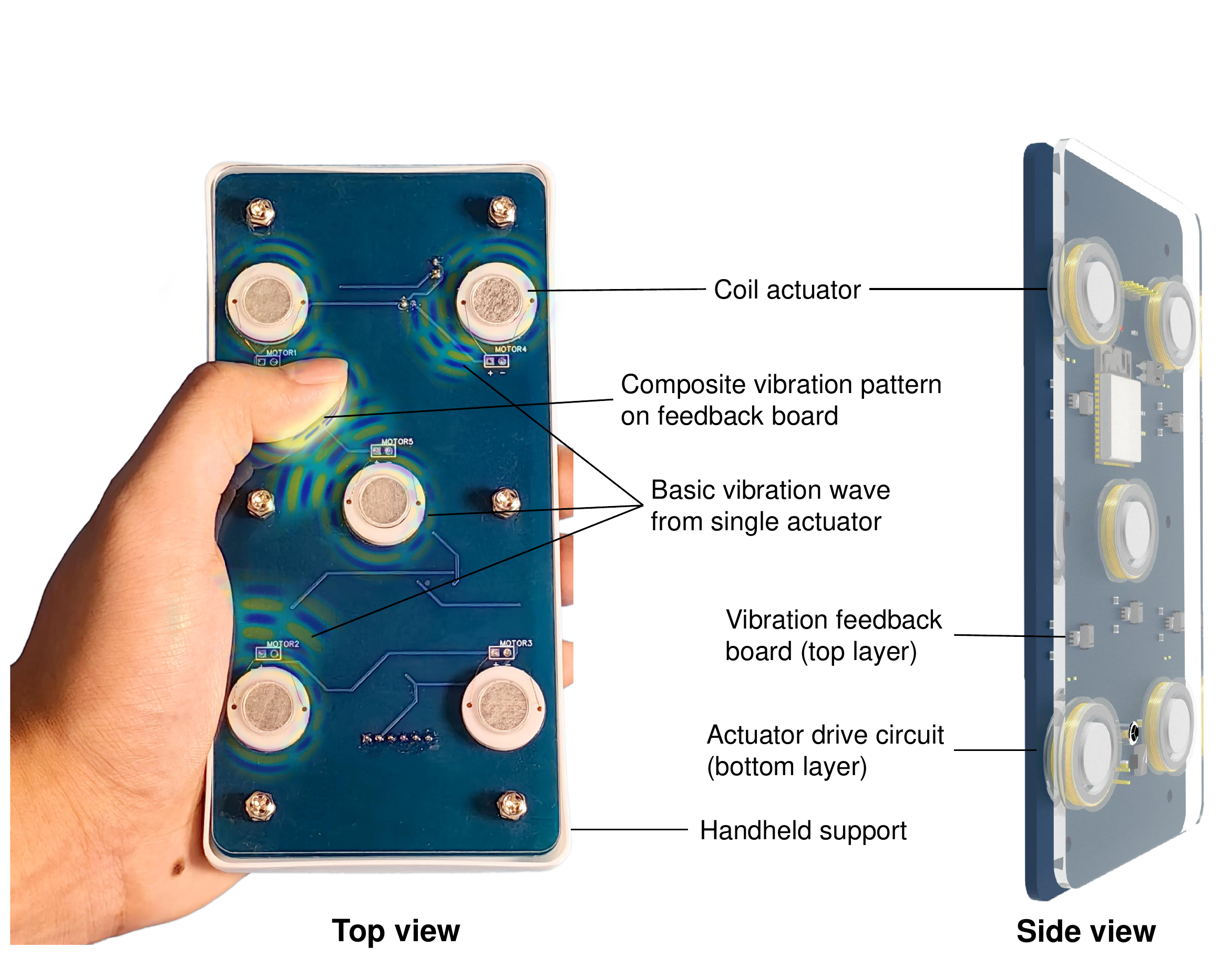}
	\caption{Point-wise vibration pattern feedback superimposed by the sparse actuator array on the top feedback board, with the bottom circuit board for actuator voltage driving.}
	\label{fig_1}
	\end{figure}

	In recent years, researchers have attempted to leverage the tactile funneling illusion~\cite{alles1970information,jones2006human} and vibration-induced location illusion\cite{dargahi2004human,lederman2009haptic}, to generate the specific vibration pattern by wave superposition using actuators with a limited amount. 
	Seo \textit{et al.}~\cite{seo2010initial} firstly investigated the tactile phantom sensation between two tactile actuators by manipulating vibration rendering, duration and direction, to create various sensations of vibration position and intensity. 
	Then they designed 32 different types of the edge-flow vibration patterns, and achieved an information transmission capacity of 3.70 bits~\cite{seo2015edge}. 
	Pantera~\cite{pantera2019sparse,pantera2020multitouch} employed the inverse filtering method to decouple multi-point vibration modes in a sparse actuator array, enabling individual fingers to feel different vibration modes. 
	To address the distortion in target vibration point within vibration patterns, Vlam \textit{et al.}～\cite{de2023focused} tried to focus the energy by utilizing the vibration wave propagation and attenuation, and achieved the generation of 9 vibration target locations within the scope of four sparse actuators. 
    These efforts on the sparse actuator array, also including~\cite{ryu2009t,kang2012smooth,basdogan2020review}, have achieved the  area-specfic vibration pattern production with a number of distinguishable species that far exceeds that of actuators. However, for any contacting point on the two-dimensional tactile feedback surface, the point-wise vibration pattern production via only a few sparse actuators still needs further research.
	
	In this study, we develop a surface tactile board based on the sparse actuator array as shown in Fig.~\ref{fig_1}, to provide point-wise vibration feedback on the contacting fingertip of handheld users. Five passive coil actuators with the same specification  are uniformly distributed on the tactile board in the size of smartphone. For single actuator with the flap-latch structure, sine wave in the frequency of 160 Hz is chosen as basic vibration wave for modulation, which owns the maximum resonant energy. According to the homogeneity and superposition of five actuator vibration patterns, we model the tactile feedback board as a Linear Time-Invariant (LTI) system with the input of single actuator vibration mode. By drive each actuator with the basic sine wave of modulated phase spectrum, vibration patterns on the board are superimposed as different energy distribution images. For the target point-wise vibration pattern production, we employ the simulated annealing algorithm to search the optimal phase spectrum for every actuator, to ensure the composite vibration energy image to have the max Structural Similarity Index Measure (SSIM) value with the target image. For random target feedback point on the board, our strategy can reach an average value of about 0.9 in the SSIM evaluation. Different point-wise vibration feedback pattern can deliver complex touching messages to the handheld users, for the immersive interaction and better tactile guidance experience.
	
	The main contributions of this article are as follows:
	\begin{itemize}
		\item For single actuator, vibration characteristic analysis with the passive flip-latch structure is elaborated by acceleration measurement, mainly about the tactile feedback board responses to varying vibration frequencies and waves for driving actuator.
		\item For the vibration feedback board, we have built an LTI system model with the input value of single actuator vibration pattern series in the timeline. This is based on the proof of the homogeneity and the superposition inherent to the vibration patterns for different actuators.
		\item For onboard vibration pattern production, point-wise vibration pattern decoupling is achieved with the random search of simulated annealing in the phase space of five actuators. By constantly getting out of local optimal solutions for actuator phase spectrum, the vibration energy distribution should finally focus on the target feedback point.
	\end{itemize}

	\section{Coil Actuator Specification }
	
\begin{figure}[htbp]
		\centering
		\includegraphics[width=8.5cm]{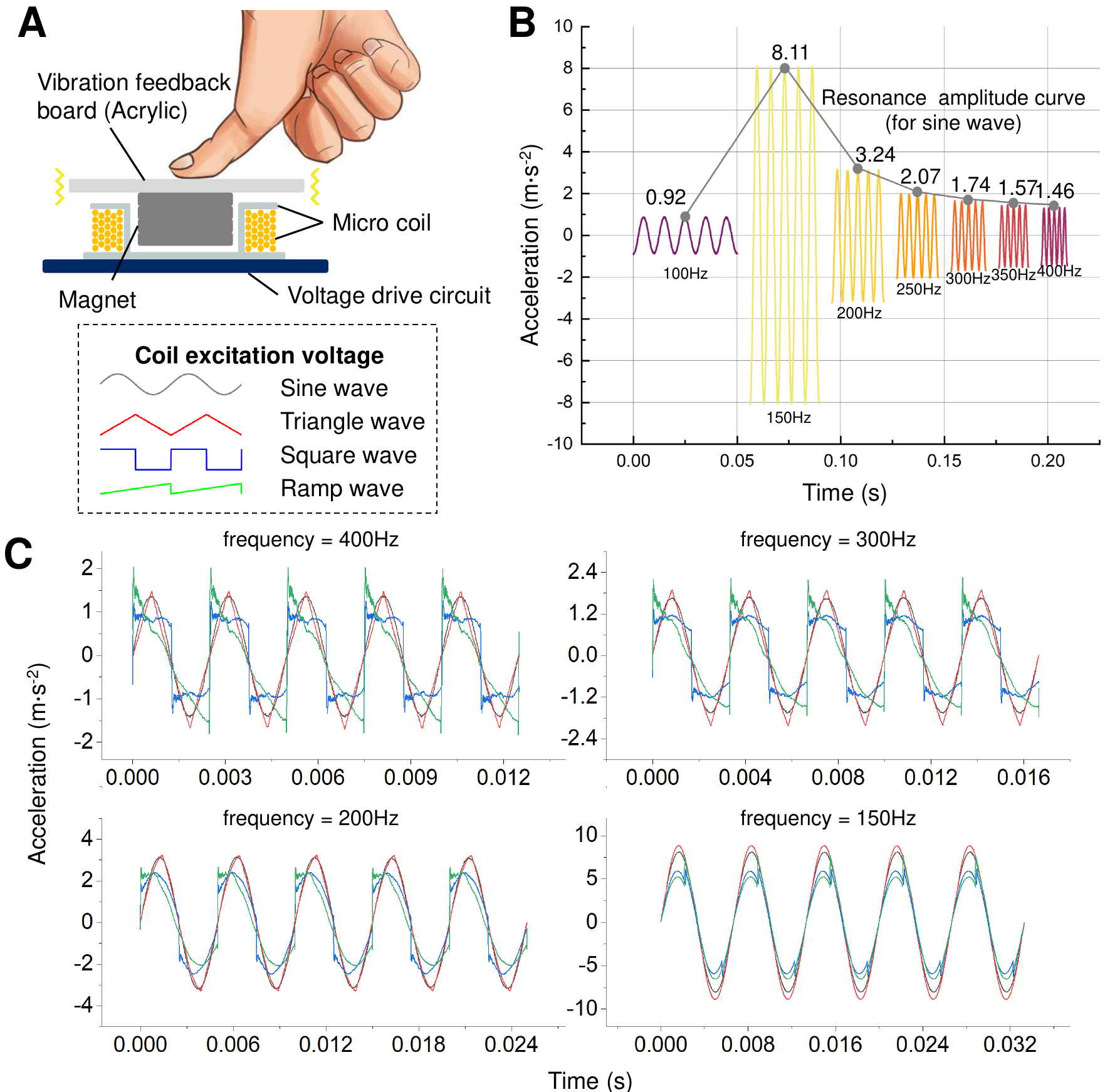}
		\caption{Specification of single coil actuator. (A) Flip-latch structure for actuator. (B) Resonance energy variation for vibration drive sine wave in the sweep frequency. (C) Time-domain vibration response in the acceleration form on the feedback board, for sine, triangle, square, and ramp wave driving in the sweep frequency, respectively.}
		\label{fig_2}
	\end{figure}

	Coil actuators serve as a crucial component for generating vibration feedback on the surface of tactile board. To facilitate full control over the vibration wave, eccentric-structure-based vibration motor with the fixed motor and frequency is eschewed, and magnet-coil assemblies utilizing a flip-latch structure is designed as depicted in Fig.~\ref{fig_2}(A). In each coil actuators, the copper wire with a diameter of 0.1 mm is wound 150 turns around an I-shaped frame, forming a hollow cylindrical coil with inner diameter of 13 mm, outer diameter of 17 mm, and height of 2 mm, which is securely affixed to the bottom current-driving circuit board. the cylindrical permanent magnet core, with a diameter of 10 mm and height of 4 mm, is situated in the center of the cooper coil, but remains non-contacting. When the coil is energized by an alternating voltage, the magnet core experiences alternating attraction and repulsion under the coil magnetic field, driving the attached acrylic tactile board in upward and downward motions to generate vibration feedback. In the latch structure, the magnet core and the coil maintain their relative positions between the top vibration board and the bottom circuit board in the static status. And the magnet core can reach a large range of vertical travel, and do incur little energy loss compared to the flexible silicone support in the typical cantilever-beam structure.
	
	We utilize a micro-controller ESP32-Wroom-32D embedded in the bottom voltage drive circuit, to output the Pulse Width Modulation (PWM) signal to emulate different alternating waves including sine wave, triangle wave, square wave, and ramp wave. And these digital voltage are then amplified via the field-effect tube FS8205a for controlling coil actuators. On the host computer, we configure the frequency and duty ratio of the original drive wave, which are then transmitted in real-time to the voltage drive board for PWM output, via the User Datagram Protocol (UDP) over the local area network. For five coil actuators as the vibration sources, simultaneous transmission of five drive waves can achieve a frequency of 10 kHz with the network latency in 10 ms. 
	
	Four representative alternating signals are selected, with frequencies ranging from 100 to 400 Hz, to serve as the drive waves for the passive coil actuator. Acceleration data at the fingertip position of the vibrating board is comprehensively collected to validate the vibratory response of single actuator to different input voltage levels. The acceleration data is acquired using a piezoelectric accelerator (SAED0005B, Wuxi Shiao) with a sampling rate of 128 kHz, a measurement range of up to 50 G, and a sensitivity of 100 mV/G. And the vibration waveform generation result is illustrated in Fig.~\ref{fig_2}(B) and (C). In terms of frequency, the vibratory response for various input waves is governed by the resonant frequency of the vibration board within the latch structure configuration. The vibration response exhibits maximal amplitude when the input waveform frequency approximates the natural frequency of the vibration board system. Taking the sine wave vibration response in Fig.~\ref{fig_2}(B) as an example, the acceleration at the test point on the vibrating panel initially increases and then decreases as the input wave frequency escalates from 100 to 400 Hz. Upon precise measurement, all four waves reach their peak vibration acceleration at 160 Hz. At this frequency, the system resonant efficiency is optimal, yielding the most robust vibration feedback with a limited input energy.
	
	Another intriguing characteristic is that around the frequency of 160 Hz, the response of all four different waves converges toward a sinusoidal shape. As illustrated in Fig.~\ref{fig_2}(C), when the frequency of the input wave significantly surpasses the resonant frequency of the vibration, the shape of the vibration response increasingly approximates the original input wave shape. And this resemblance becomes even more pronounced as the frequency increases. As the frequency of input wave nears the resonant frequency, the vibration response of wave other than the sine wave also starts approximating a sinusoidal shape. This convergence reaches its peak at 160 Hz. Even so, the response to sine-wave input retains its maximum amplitude, while the amplitude of other waves experiences a certain degree of energy loss compared to the sine wave. Consequently, we can conclude that when driven by the 160 Hz sine wave, single coil actuator in our tactile board exhibits the optimal generation efficiency of vibration feedback.

	\section{Vibration Board Modeling}
	The tactile feedback board employed in this article is fabricated from a 1 m-thin acrylic sheet with dimensions 71.5$\times$146.7 mm$^2$. When vibration generating, the permanent magnet within the coil actuator is subjected to the coil alternating magnetic field, causing vertical motion that induces bending deformations at the corresponding locations on the thin elastic acrylic board. At a actuator vibration frequency of $f$(=160 Hz), the bending wavelength of the board vibration is approximately 187 mm, substantially greater than its thickness. Therefore, we can qualify the plate as an isotropic and homogeneously thin elastic plate.
	
	\subsection{Vibration Pattern of Single Actuator}
	Note the position of single actuator on the vibration board as $(x_0, y_0)$, and the drive wave amplitude for vibration induced on the thin plate is $F(t) = A\cdot$ sin $(2\pi ft + \phi)$. Considering that the displacement is $w(x, y, t)$ at any observation position $(x, y)$ on the vibration feedback board at any time $t$ within the board boundaries $(0 \leq x\leq a, 0 \leq y \leq b)$. The board's motion can be described by the dynamic Kirchhoff plate model, taking into account the plate's bending stiffness, which is influenced by both its own mass and external loading. The governing equation for the motion of the whole vibration board is represented as:
	\begin{equation}
		D\nabla^4 w + \rho h \frac{\partial^2 w}{\partial t^2} = F(t)\delta(x - x_0)\delta(y - y_0),
		\label{eqn_1}
	\end{equation}
	where $D = Eh^3/(12(1-\nu^2))$
	denotes the flexural rigidity of the plate, a function about the thickness $ h $, Young's modulus $E$, and Poisson's ratio $\nu$. Here, $\nabla^4$ is a bi-harmonic operator, and $ D\nabla^4 w$ represents the plate's restoring force under the influence of external loads. $\rho$ and $h$ are the density and thickness of the plate respectively, and $\frac{\partial^2 w}{\partial t^2}$ reflects the plate's inertia. $\delta$ is the Dirac delta function, implying the external load only exists at the actuator position $(x_0,y_0)$.
	
	Throughout the vibration process, each mode of the vibration feedback board has its specific spatial distribution and time dependency. Their product indicates the contribution of this spatio-temporal mode to the overall displacement of the board. The displacement $w(x,y,t)$ at any point and time can be viewed as the superposition of all spatio-temporal modes. Based on the properties of the partial differential equation, the displacement can be written as an eigenfunction expansion in the modal space:
	\begin{equation}
		w(x, y, t) = \sum_{{m=1}}^{\infty} \sum_{{n=1}}^{\infty} \phi_{mn}(x, y) q_{mn}(t).
		\label{eqn_2}
	\end{equation}
	Here, $ m $ and $ n $ serve as the modal coordinates. $\phi_{mn}(x, y)$ is the eigenfunction representing the spatial distribution (modal shape function) of mode $(m, n)$ for the feedback board. And $q_{mn}(t)$ denotes the time-dependent coefficient representing the amplitude variation of the corresponding mode at the corresponding time. 
	Substituting this expansion back into the original Eqn.~\ref{eqn_1} governing the vibration of the thin board, and taking into account the orthogonality of the eigenfunctions, then the normalized eigenfunctions $ \phi_{mn}(x, y) $ can be computed as follows:
	 \begin{equation}
		 	\phi_{mn}(x, y) = \frac{2}{\sqrt{ab\rho h}} \sin\left(\frac{m\pi x}{a}\right) \sin\left(\frac{n\pi y}{b}\right).
		 \end{equation}

	 Given the eigenfunctions of the feedback board, the time-dependent coefficients $\Omega_{mn}(t)$ corresponding to these eigenfunctions can be calculated through integration with respect to the external load $F(t)$:
	 \begin{equation}
		 	\Omega_{mn}(t) = \int_0^b \int_0^a F(t) \cdot \phi_{mn}(x, y) \, dx \, dy.
		 \end{equation}
		 
	 In the context of the temporal dependency coefficient $ q_{mn}(0) $
	 pertaining to the vibration feedback plate, it is worth noting that the initial displacement of the plate is zero, leading to the initial conditions 
	 $q_{mn}(0) = \frac{\partial q_{mn}(0)}{\partial t} = 0$.
	 When the external load $F(t)$ manifests as a sinusoidal wave and is independent of the spatial coordinates, its eigenfunction dependency $\Omega$ is also sinusoidal in nature, succinctly expressed as $\Omega_{mn}(t) = \Omega_0 \cdot A \sin(2\pi f t + \phi)$ with a fixed value  $\Omega_0$. Under these circumstances, let $\lambda$ denote the eigenvalue corresponding to the eigenfunction. The particular solution of 	$q_{mn}(t)$ will be a sinusoidal wave that possesses the same frequency as the external load,
	 \begin{equation}
		 	q_{mn}(t) = \frac{\Omega_0}{D\lambda_{mn}^2 - \rho h (2\pi f)^2} \cdot A \sin(2\pi f t + \phi).
		 \end{equation}
	
\begin{figure}[htbp]
		\centering
		\includegraphics[width=8.5cm]{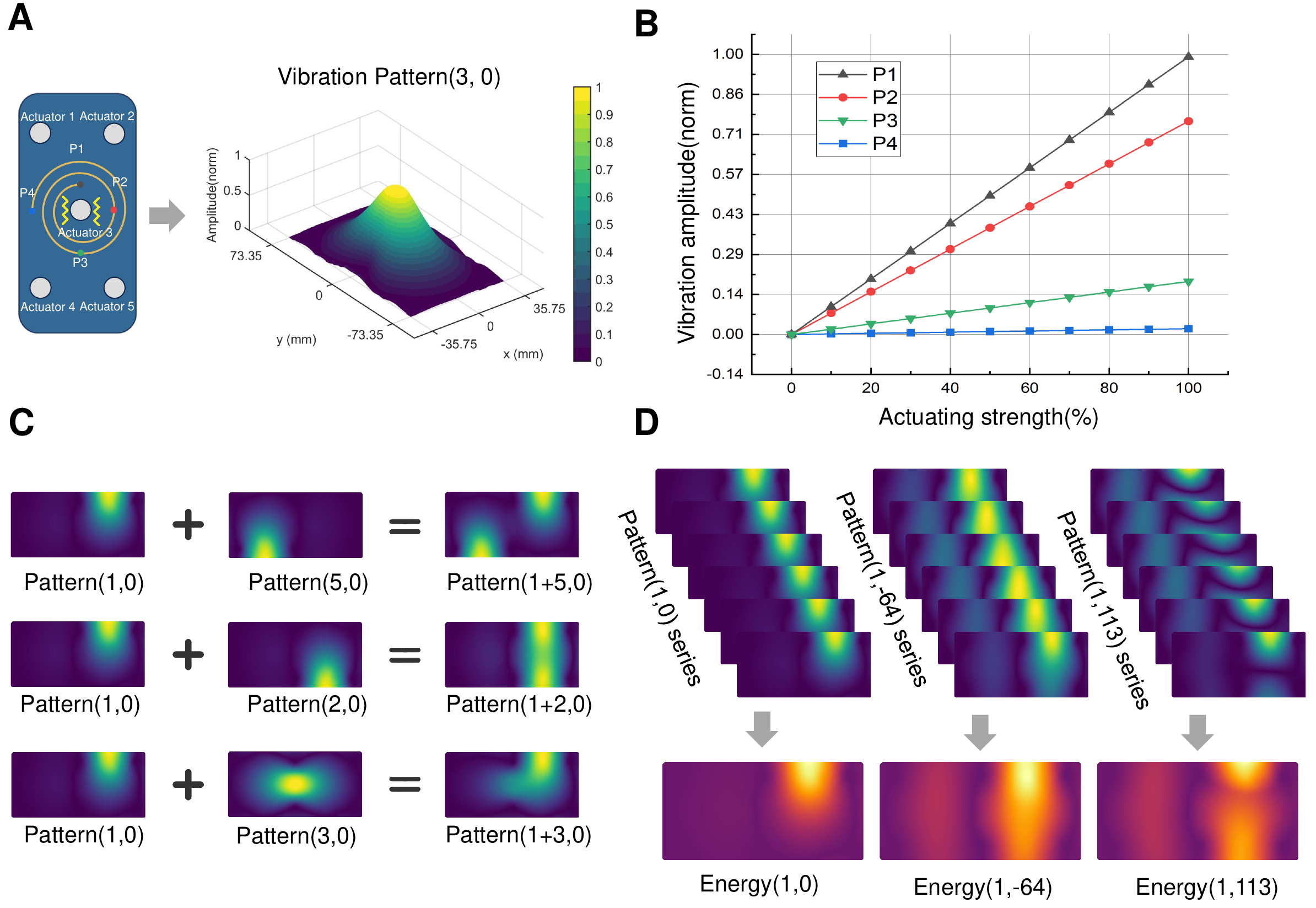}
		\caption{Modeling of the vibration feedback board system. (A) Vibration pattern generation for single actuator. (B) The linearity between the onboard vibration amplitude and the actuator driving strength. (C) Superposition of vibration pattern form different actuators. (D) Transfer the vibration patterns in single period to the energy distribution image, for satisfying the human tactile perception ability.}
		\label{fig_3}
	\end{figure}

	To obtain the spatial vertical displacement response $w(x, y, t)$ of the vibration feedback board at any given time $t$, we need to sum over the terms $\phi_{mn}(x, y)$ and $q_{mn}(t)$ in the space and time domain. By solving Eqn.~(\ref{eqn_2}), we can generate a graphical representation of the vibration state of the entire feedback plate at arbitrary time instants. As an example, the vibration pattern at a specific time instance within a stable periodic cycle for the middle Actuator 3 with a phase of 0 is depicted in Fig.~\ref{fig_3}(A) at a resolution of 179$\times$367 points.
	
	\subsection{Homogeneity and Superposition of Multiple Patterns}
	In the aforementioned comprehensive solution process, the time-dependent coefficient $ q_{mn}(t) $ exhibiting a linear relationship to the drive wave amplitude $F(t)$ under the sinusoidal shape. And this homogeneity remains between the amplitude of the vibration response $ w(x, y, t) $ and the drive wave. This linear relationship was empirically verified by measuring the amplitudes at various observation points located at different distances and directions around Actuator 3, as shown in Fig.~\ref{fig_3}(B). Within the range of vibration intensities that can be applied by the vibration board, a consistent linear correlation has been maintained between the amplitude of the actuator driving wave and the vibration feedback. For any single actuator $i \in \{1, 2, 3, 4, 5\}$, the vibration pattern $\mathbf{P}$ under the driving wave of amplitude $A$ and phase $\phi$ is represented, in light of this linearity, as $A \cdot \mathbf{P}(i, \phi)$.
	
	The vibrating plate used in this study is made of uniform acrylic material, and the vibration patterns induced by different coil actuators can be directly summed at any given time instant to produce a composite vibration pattern. We have validated this superposition relationship of vibration patterns originating from different actuators, as demonstrated in Fig.~\ref{fig_3}(C). By independently driving Actuator 1 and its opposite, diagonal, and central actuators, their time-sequential vibration patterns is numerically summed within a stable cycle. The result is found to be entirely consistent with the time-sequential vibration patterns obtained by two simultaneously-driving actuators. 
	In rigorous mathematical terms, the composite vibration profile, generated by the simultaneous operation of multiple actuators $i_1, i_2, \ldots, i_n$ with amplitudes $A_1, A_2, \ldots, A_n$ and phases $\phi_1, \phi_2, \ldots, \phi_n$, can be expressed as the linear combination of their individual contributions.
	Specifically, we denote this property of additive superposition of vibration patterns from disparate sources, along with the linearity, as the formula:
	\begin{equation}
		\begin{aligned}
			\boldsymbol{A} \cdot \mathbf{P}([i_1, i_2, \ldots, i_n], [\phi_1, \phi_2, \ldots, \phi_n]) &= \hfill \\ 
			A_1 \! \cdot \! \mathbf{P}(i_1, \phi_1) \! + \!  A_2 \cdot \mathbf{P}(i_2, \phi_2) \! + \! \cdots\! +\! & A_n\! \cdot\! \mathbf{P}(i_n, \phi_n).
		\end{aligned}
		\label{eqn_3}
	\end{equation}
	
	Human vibration perception exhibits a specific sensitivity across different dimensions such as frequency, amplitude, and waveform. The perception is particularly acute within the frequency range of 20$\sim$500 Hz. However, for the 160 Hz high-frequency sinusoidal wave employed on this vibration board,  it is challenging to discern waveform variations in the time domain resulting from phase modulation. Under this circumstance, the sensation induced by the 160 Hz vibration resembles fluctuations in a form of energy rather than the wave shape. For the single-period vibration pattern series, the Root Mean Square (RMS) of the vibration amplitude at each point in the spatial frame is calculated over time, serving as a measure as the vibration energy distribution image with the same size of vibration pattern. The vibration energy distribution $\mathbf{E}$ across the entire tactile feedback board can be represented as follows:
	\begin{equation}
		\mathbf{E} = \text{RMS}(\mathbf{P}).
		\label{eqn_4}
	\end{equation}
	Accordingly, vibration pattern sequences can be transformed into energy images that better quantify the human tactile perception characteristic, as shown in Fig.~\ref{fig_3}(D). Even unlike the original vibration patterns with  both positive and negative value, vibration energy images lack polarity but retain high similarity with the original patterns.
	
	Besides, vibration energy image elucidates the clear impact of  drive wave phase on the final vibration pattern. Even for the same actuator, different phases of the sinusoidal driving waveform result in distinct vibration energy images due to differing initial conditions. This diversity enriches the basic vibration modes emanating from a sparse array of actuators, enabling the superposition of targeted single-point vibration pattern feedback by modulating the phase spectrum of each actuator.

	\section{Vibration Pattern Decoupling}
	
	When vibration waves originating from five actuators propagate from disparate locations, and converge at a specific point on the vibration feedback board, their induced longitudinal displacements sum up in accordance with the principle of linear superposition. And this superposition process is reversible for the LTI vibration system. It implies that for a vibration distribution image with an arbitrary target point as the source, it is possible to decouple the contributions from vibration waves emanating from a sparse array of actuators.
	
	The generation of the target vibration energy distribution, $\mathbf{E}_T$, serves as the initial step in the entire process of vibration pattern decoupling. To ensure the reversibility of the vibration synthesis and decoupling process, the target energy distribution must be derived from the vibration patterns of single-point vibration source. For any target vibration feedback point $\boldsymbol{T}(x_T, y_T)$, Eqn.~(\ref{eqn_2}) is employed to solve for the vibration patterns of single actuator at position $\boldsymbol{T}$. Then, the target vibration energy distribution, $\mathbf{E}_T$, is obtained by taking the RMS value of the vibration energy at each point across the vibration feedback board by Eqn.~(\ref{eqn_4}).
	
	During the process of vibration decoupling, we select unit-amplitude vibration pattern sequences from five actuators, each at all 360 phases by one degree, as the basic vibration modes. By adjusting each basic vibration mode with the respective amplitude (constrained to be within ten times the unit amplitude), we synthesize the vibration pattern sequences across the entire feedback board according to Eqn.~(\ref{eqn_3}). The final objective of the decoupling process is to make the energy distribution $\mathbf{E}$, of the synthesized vibration pattern , as similar as possible to the target energy distribution $\mathbf{E}_T$. Here, we choose the SSIM as the evaluation metric, with a particular focus on the contrast of vibration energy on different areas. Within the range from 0 to 1, the closer the SSIM value between the synthesized and target vibration energy distributions approaches 1, the more effective is the generated vibration pattern effect.
	
	\begin{figure}[htbp]
		\centering
		\includegraphics[width=8.5cm]{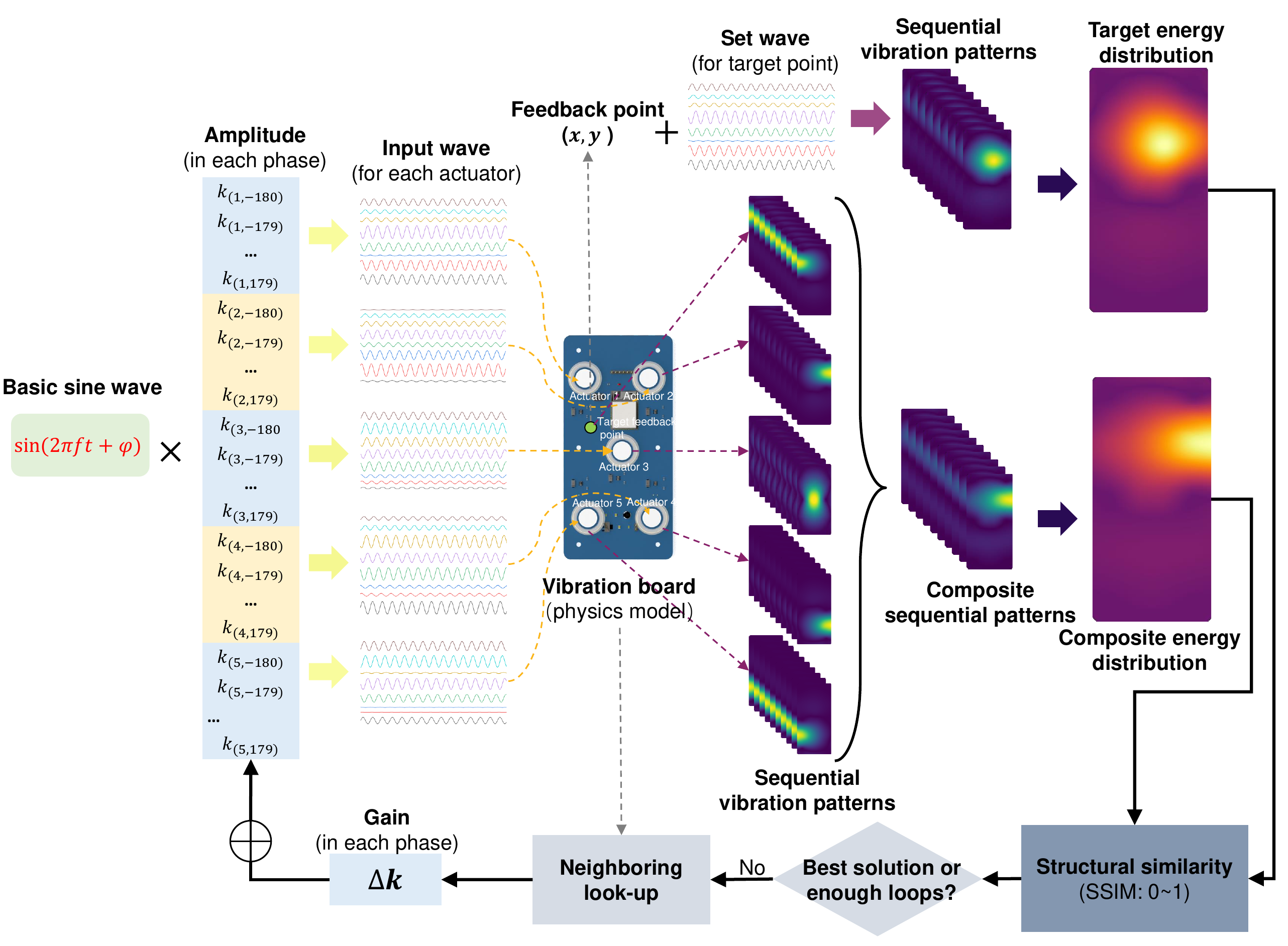}
		\caption{The global optimal amplitude spectrum searching in all phases for each actuator, based on the vibration superposition model and the simulated annealing algorithm.}
		\label{fig_4}
	\end{figure}
	
	For the vibration pattern decoupling, the efficacy evaluation is based on the energy distribution image, whereas the vibration synthesis process is reliant on the sequence of vibration patterns. Consequently, this decoupling problem is not amenable to gradient descent or machine learning-based optimization methods. During the application of stochastic search algorithms, it is crucial to consider that the vibration decoupling problem might possess multiple optimal solutions, necessitating a strategy to avoid the entrapment in local optima. For the synthesized and target energy images, their SSIM value can be construed as a form of energy. Throughout the optimization process, the similarity value incrementally ascends from a lower level to achieve the stabilization at a high level, allowing for the possibility of escaping local optima through an approach analogous to simulated annealing. Specifically, when the SSIM value is low, the optimization process is configured to accept bad solutions at the current step with a relatively high probability. The entire vibration energy distribution decoupling process based on the simulated annealing approach is detailed in \textbf{Algorithm}~\ref{alg_1} and Fig.~\ref{fig_4}.
	
	\begin{algorithm}
		\caption{Vibration Energy Distribution Decoupling.}
		\label{alg_1}
		\KwIn{Basic vibration pattern series for 5 coil actuators in the 360 phases: $\mathbf{P}(i, j)$; Target vibration feedback point:$\left(x_T, y_T\right)$.}
		\KwOut{Amplitude gain coefficients for each coil actuator in its all phases: $\boldsymbol{k}(i, j)$, which makes the composite vibration pattern closest to the target pattern.}  
		\BlankLine
		Simulate the target vibration energy distribution $\mathbf{E}_T$;\\
		Initialize all the gain coefficients  $\boldsymbol{k}_0(i, j)$ randomly;\\
		Get the energy distribution of initial composite patterns $\mathbf{E}_0\left(\boldsymbol{k}_0 * \mathbf{P}\right)$;\\
		\While{\textnormal{$n$ \textbf{in} range(10000)}}{
			Generate the all the gain offset $\boldsymbol{k}_n=\boldsymbol{k}_{n-1}+\Delta \boldsymbol{k}$;\\
			Add the basic vibration patterns with gain $\boldsymbol{k}_n * \mathbf{P}$; \\
			Get the energy of composite patterns $\mathbf{E}_n\left(\boldsymbol{k}_n * \mathbf{P}\right)$;\\
			Compute the energy distribution similarity $\operatorname{SSIM}\left(\mathbf{E}_n, \mathbf{E}_0\right)$;\\
			\If{$\operatorname{SSIM}\left(\mathbf{E}_n, \mathbf{E}_0\right)<\operatorname{SSIM}\left(\mathbf{E}_{n-1}, \mathbf{E}_0\right)$ \textnormal{\textbf{or}} $1-\operatorname{SSIM}\left(\mathbf{E}_{n-1}, \mathbf{E}_0\right)<\operatorname{rand}(0,1)$}{
				Accept this new gain offset of $\Delta \boldsymbol{k}$;\\
			}
			\Else{
				Do not update the gain $\boldsymbol{k}_n=\boldsymbol{k}_{n-1}$;\\
			}
		}
		Output the locally optimal solution for amplitude gain coefficients $\boldsymbol{k}(i, j)=\boldsymbol{k}_{\text {10000}}$
	\end{algorithm}

	\section{Experiments and Results}
	
	During the experiment process, vibration patterns at several different position were generated and calculated as the target energy distributions. Vibration pattern sequences originating each of 5 actuators for all 360 phases within single period, were stored as fundamental vibration modes. Leveraging the homogeneity and superposition properties of our feedback board, these fundamental vibration modes were linearly combined using the optimal phase spectrum for actuators obtained through the vibration decoupling process, and then a synthesized vibration energy image that approximates the target single-point vibration feedback was obtained.
	
	\begin{figure}[htbp]
		\centering
		\includegraphics[width=8.5cm]{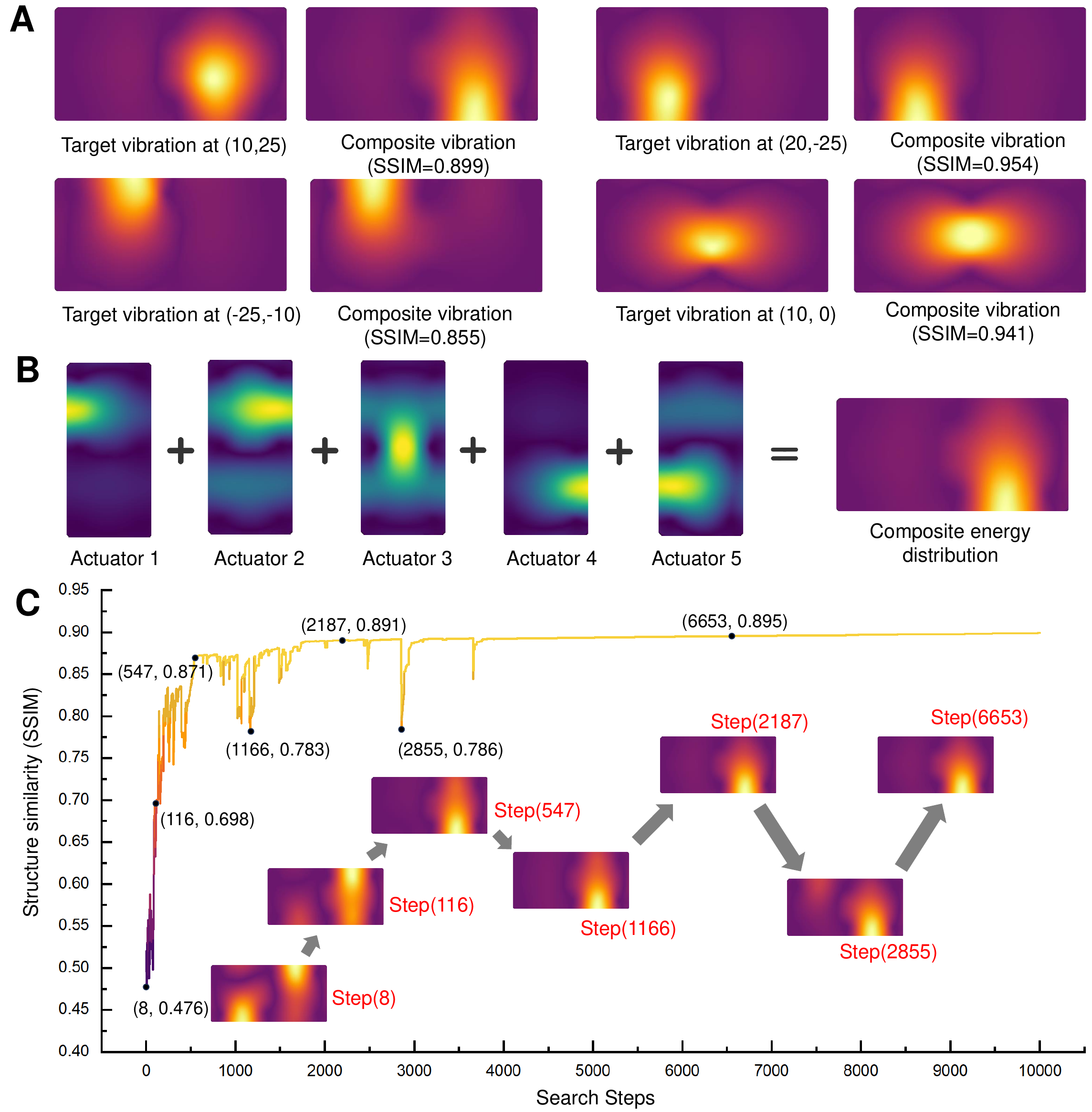}
		\caption{Effect of point-wise vibration pattern generation on the tactile feedback board. (A) Comparison between composite vibration energy pattern and target pattern for different target feedback points. (B) Raw vibration patterns from five actuators are superimposed as the optimal energy distribution image closet to the target pattern. (C) SSIM value promotion of vibration energy distribution during the simulated annealing search process. }
		\label{fig_5}
	\end{figure}
	
	The effectiveness of generating point-wise vibration image on the tactile feedback board is illustrated in Fig.~\ref{fig_5}(A). Taking four random target feedback points as examples, the SSIM value between the synthesized vibration image obtained through vibration decoupling and the target single-point vibration image ranged between 0.85 and 0.95. In each synthesized vibration image, the focal point of energy concentration in the synthesized vibration feedback closely approximated the target vibration feedback point, without any energy concentration in other areas. The localized composite vibration energy distribution demonstrates the superiority of the vibration decoupling method proposed in this study. And the target vibration energy distribution for the first target point (10, 25), achieved through the superposition of different vibration patterns emanating from five actuators, is shown in Fig.~\ref{fig_5}(B).
	
	Taking the vibration image decoupling at the target point (10, 25) as an example, we plotted the process of simulated annealing to search for the closest vibration energy image in Fig.~\ref{fig_5}(C), which was able to be completed in 8$\sim$9 hours. A randomly initialized phase spectrum was selected as the initial solution, where the SSIM value was 0.476 at Step(8). As the search process progressed, energy concentration areas far from the target feedback point gradually disappeared, and the SSIM index increased to 0.698 at Step(116). Then, by overcoming several local optimal values, the only energy concentration point progressively approached the target position, and the SSIM similarity index had risen to 0.871 by Step(547). In the subsequent search process, the optimal synthesized energy image had essentially stabilized, with the SSIM value remaining around 0.9 due to minor parameter adjustments in the search process. At this point, the algorithm had approximated the global optimum solution, achieving the maximum similarity between the synthesized vibration image and the target point vibration image. And point-wise vibration pattern production at any location of the tactile board surface was able to be achieved through this vibration decoupling process.

	\section{Conclusion}
	In this paper, we achieve the arbitrary single-point vibration feedback production on a two-dimensional tactile board through the phase modulation of driving waves for a sparse actuator array. Firstly, for single coil actuator with a flip-latch structure, we have measures its vibration responses to different driving wave, and find that the 160 Hz sine wave with the maximum resonance energy is best worked as the basic wave for subsequent vibration drive waveform modulation. Then, for our vibration feedback board with five actuators, we derive the onboard vibration patterns for single actuator and established the LTI system model for the vibration pattern synthesis  based on the verified homogeneity and superposition. Finally, we applied the simulated annealing algorithm for random search, to find the optimal phase spectrum for each actuator's vibration drive wave. The optimal phase spectrum is used to modulate the basic sine wave, so that the vibration energy image concentrated near the target vibration feedback point can be generated, with the maximum SSIM value to the target image energy. 
	
	As a conclusion, when generating the point-wise vibration image, we exclude modulation parameters with poor effects like basic wave forms and frequencies, through the physical measurement of single actuator vibration generation. The establishment of linear vibration pattern synthesis model simplifies the global random phase spectrum search, allowing the search process for the phase spectrum with the highest vibration image similarity to be completed within a few hours. For the vibration feedback generation focused on any point, the optimal phase spectrum corresponding to this position can be pre-stored in a look-up table. By judging the contact point on handheld touch devices, the actuators for vibration display can quickly concentrate the vibration energy at the target location, thus achieving the effect of arbitrary position vibration feedback with a sparse actuator array.
	
	\bibliographystyle{IEEEtran}
	\balance
	\bibliography{bib/ref}

\end{document}